# A Face-like Structure Detection on Planet and Satellite Surfaces using Image Processing

Kazutaka Kurihara[1], Masakazu Takasu[2], Kazuhiro Sasao[3], Hal Seki[4],
Takayuki Narabu[5], Mitsuo Yamamoto[6], Satoshi Iida[7], Hiroyuki Yamamoto[8]

1 National Institute of Advanced Industrial Sci. and Tech., `k-kurihara@aist.go.jp`,
2 teamLab. INC, `takasu@team-lab.com`,
3 Nico-TECH, `k.sasao@gmail.com`,
4 Georepublic Japan, `hal@georepublic.de`,
5 NTT DATA CCS CORPORATION, `tnarabu@nttdata-ccs.co.jp`
6 Denso IT Laboratory, Inc., `miyamamoto@d-itlab.co.jp`,
7 Nico-TECH, `nyampire@gmail.com`
8 DENSO CORPORATION, `hyamamyto1981@gmail.com`

**Abstract.** This paper demonstrates that face-like structures are everywhere, and can be detected automatically even with computers. Huge amount of satellite images of the Earth, the Moon, the Mars are explored and many interesting face-like structure are detected. Throughout this fact, we believe that science and technologies can alert people not to easily become an occultist.

## 1 Introduction

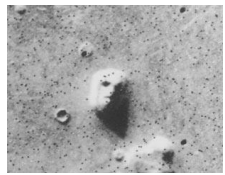

**Fig. 1.** Face on Mars ©Google

"Face-like" structures have been found in many provinces on the earth or other stars [5][2]. They are rocks or other geographical structures that look like a human face. Because it is quite rare for such shapes to be made without artificial efforts, people tend to think that they derived from such as unidentified ancient civilizations or space aliens.

One of the most popular examples of face-like structures is "Face on Mars" (Figure 1) [5]. Some people say that is an ancient ruin made by Martians, and other people say that is an artifact made by human beings. This mysterious face has been inspired many sci-fi stories [1][3], even after it turned out just an optical illusion by later analyses in detail.

Our goal of this research is to demonstrate that such face-like structures are everywhere, and can be detected automatically even with computers. Throughout this fact, we believe that science and technologies can alert people not to easily become an occultist.

## 2      Related Work

Onformative[1] shows a similar idea in 2013, but our project is original as the upload date of a Youtube video[2] shows that we demonstrated our early result in 2010.

## 3      System Description

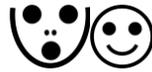

**Fig. 2.** Positive examples for training

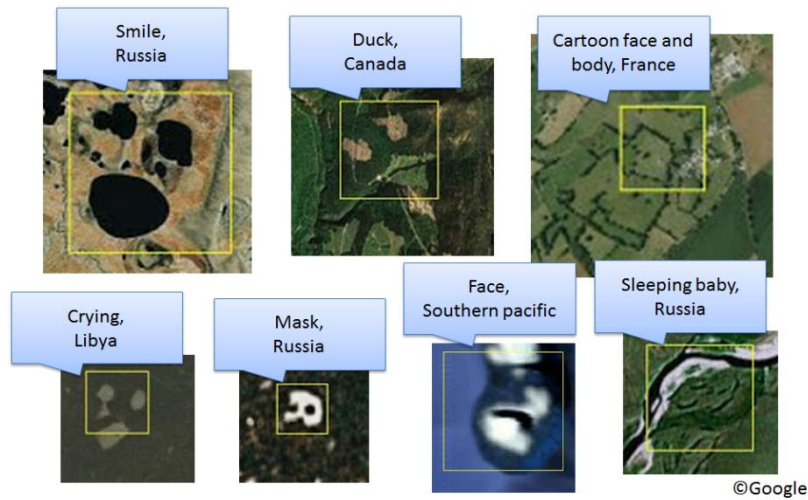

**Fig. 3.** Example results ©Google

### 3.1    Prototype #1 for the Earth Surface

The object detector of OpenCV with Haar-like features was used to detect face-like structures. This was a commonly used classifier to detect human faces. Since features of face-like structures such as Figure 1 were slightly different from those of original human faces, we have trained our classifier with some simplified illustrations of human faces like Figure 2 as positive examples.

---

[1] http://www.onformative.com/lab/googlefaces/
[2] https://www.youtube.com/watch?v=ooKFEmE6JXQ

In 2010, we collected thousands of satellite images per day using Google Static Map API [4] and applied our classifier[3]. Figure 3 shows some examples of our results.

With this prototype, we learned that our detection result included many false positives: face-like clouds, and simple mis-recognitions that were not recognized as human-faces with human cognition, and need some improvements to refine detection algorithm.

### 3.2 Prototype #2 for the Mars and the Surfaces

We developed our prototype #2 as a project for International Space App Challenge hosted by NASA[4]. The major functional updates are (1) applying multiple detection algorithms in parallel, and (2) applying the detector to the Mars and the Moon surfaces that has less obstacles such as clouds.

**Applying Multiple Detection Algorithms in Parallel**

We applied three algorithms in parallel to our face detector: OpenCV Haar-like feature detection, Anime face detection[6], and Brightness Binary Feature[7][8]. 50 instances of Windows Asure were used for the computation.

**Applying Detector to Mars and Moon Surfaces**

We explored face-like structures from 11,829,761,106 pixels, including 10,059,979,678 pixels on the Mars surface, and 533,491,975 pixels on the Moon surface, and 1,236,289,453 pixels of others, provided by NASA Space App Challenge.

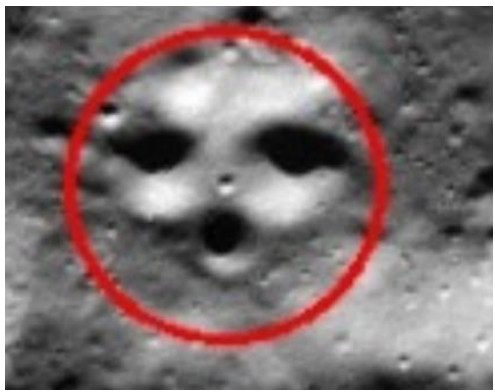

**Fig. 4.** "Grey Aliens" on the Moon ©NASA

---

[3] https://sites.google.com/site/geofaceproject/
[4] We won the third place award in the Tokyo local competition. http://marproject.org/

**Results**

Figure 4 shows a detection example that looks like a famous alien called "Grey Aliens" [9]. This structure is located on the south pole of the moon [10]. We could find many face-like structures in the polar area because the incidence angle of sunlight was smaller in high latitude and inside of craters were mostly dark with shadows. Other examples of faces on the south and north poles of the moon were shown in Figure 5.

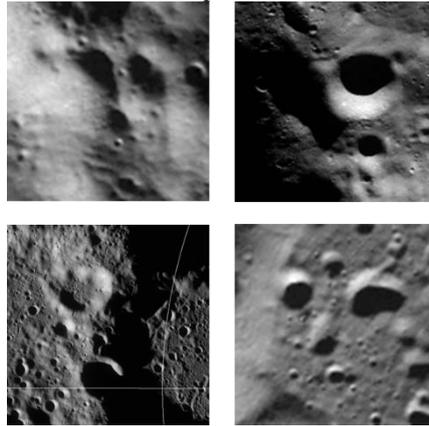

**Fig. 5.** Face-like structures on the south and north pole of the Moon.

## 4    Conclusion and Future Work

In this paper we demonstrated that face-like structures are everywhere and can be detected automatically on varieties of star surfaces, even with computers. Out next step is to manage our web service that allows people to vote "face candidates" to refine automatic detection results and obtain better and interesting face-like structures with collaborative intelligence.